\documentclass[runningheads]{llncs}
\usepackage{amsmath,amssymb}
\usepackage{mathtools}
\usepackage{graphicx}
\usepackage{subfig}
\usepackage{afterpage}
\usepackage{multirow}

\newcommand{\uncertaintyfactor}{\ensuremath{\rho}}
\newcommand{\sdt}[3]{\ensuremath{\texttt{SDT}_{#1}\left[#2\right](#3)}}
\newcommand{\mcsdt}[3]{\ensuremath{\texttt{MC-SDT}_{#1}\left[#2\right](#3)}}
\newcommand{\detsdt}[3]{\ensuremath{\texttt{DET-SDT}_{#1}\left[#2\right](#3)}}

\newcommand{\referenceset}{\ensuremath{X}}
\DeclareMathOperator{\EX}{\mathbb{E}}
\newcommand{\dmax}{\ensuremath{d_{\texttt{MAX}}}}
\newcommand{\prob}[1]{\ensuremath{\mathbb{P}(#1)}}
\newcommand{\deuclidean}{\ensuremath{d_{E}}}

\newcommand{\randomize}{R}
\newcommand{\drs}{\ensuremath{Y}}
\newcommand{\darrow}{\ensuremath{d_{\rightarrow}}}
\newcommand{\rpm}{\raisebox{.2ex}{$\scriptstyle\pm$}}
\newcommand{\Tstrut}[1][2.6ex]{\rule{0pt}{#1}}

\begin{document}
\title{Stochastic Distance Transform}
%
%
\author{Johan \"{O}fverstedt\inst{1} \and
Joakim Lindblad\inst{1,2} \and
Nata\v{s}a Sladoje\inst{1,2}}
\authorrunning{J. \"{O}fverstedt et al.}
%
\institute{$^1$ Centre for Image Analysis, Uppsala University, Sweden \\
$^2$ Mathematical Institute of the Serbian Academy of Sciences and Arts, Serbia\\
\email{johan.ofverstedt@it.uu.se}, \email{joakim@cb.uu.se}, \email{natasa.sladoje@it.uu.se}\\
}

\maketitle              
\begin{abstract}
The distance transform (DT) and its many variations are ubiquitous tools for image processing and analysis.
In many imaging scenarios, the images of interest are corrupted by noise. This has a strong negative impact on the accuracy of the DT, which is highly sensitive to spurious noise points. In this study, we consider images represented as discrete random sets and observe statistics of DT computed on such representations. We, thus, define a {\it stochastic distance transform} (SDT), which has an adjustable robustness to noise. Both a stochastic Monte Carlo method and a deterministic method for computing the SDT are proposed and compared. Through a series of empirical tests, we demonstrate that the SDT is effective not only in improving the accuracy of the computed distances in the presence of noise, but also in improving the performance of template matching and watershed segmentation of partially overlapping objects, which are examples of typical applications where DTs are utilized.

\end{abstract}
%
%

\section{Introduction}

Distance transforms (DTs) have been, since introduced to image analysis in 1966,
\cite{rosenfeld1966sequential},
a standard tool with applications in, among others, the context of similarity measure computation \cite{lindblad2009set}, image registration and template matching \cite{lindblad2009set}, segmentation \cite{beucher1979use}, and skeletonization (by computing the centres of maximal balls) of binary images \cite{saha2016survey}.


The properties of DT have been explored extensively, and several aspects of their performance have been improved by a sequence of important studies:  their optimization for efficient approximation of Euclidean DT by local computations \cite{borgefors1986distance},  fast algorithms for the exact Euclidean DT \cite{maurer2003linear}, DT with sub-pixel precision \cite{gustavson2011anti,lindblad2015exact}, and extension of DT to grey-scale and fuzzy images \cite{levi1970grey,saha2002fuzzy}. Exact algorithms eliminate approximation errors, and sub-pixel precision methods can reduce the inaccuracy of distances introduced by digitization of objects. However, neither provide a solution to the challenge which spurious points and structures pose: a single noise-point can be sufficient to heavily corrupt the DT, and negatively affect performance of all the analysis methods relying on it. 

In this study we combine prior theoretical work related to discrete random sets (DRS) with the approach to observe distributions over distances, and we propose a Stochastic Distance Transform (SDT). Rather than increasing precision, SDT introduces a gradual (adjustable) insensitivity to noise, acting to yield a smoothed DT with less weight attributed to sparse points. A stochastic Monte Carlo-based method and a deterministic method are proposed for computing the SDT. 
%
A similar idea is explored in \cite{moreno2013robust}, where a robust method for estimating distributions of Hausdorff set distances between sets of points, based on random removal of the points in the observed sets, is proposed. In that work, the authors utilize DT only as a tool for estimation of the Hausdorff set distance by computing weighted distance histograms based on user-provided point-wise reliability coefficients, without exploring how these random sets can increase the robustness and accuracy of the DT itself. We fill this gap and, in this study, define and evaluate the SDT, with confidence that it can find applications in numerous other scenarios where DTs are used.   

We show that the proposed method is more accurate than the standard DT in the presence of noise, and that it can increase the performance of several common applications of the DT, such as template matching and watershed segmentation.

\section{Background}

\subsection{Discrete Random Sets}

Discrete random sets (DRS) \cite{matheron1975random,goutsias1992morphological} are random variables taking values as subsets of some discrete reference space. DRS theory provides a theoretical foundation and offers suitable tools for the modeling and analysis of shapes in images, allowing exploration of both their structural and statistical characteristics.
Representing (binary) image objects as (finite) random subsets on an image domain (bounded on $\mathbb{Z}^n$) facilitates their structural analysis in presence of noise.


The coverage probability function of a DRS $\drs$ is defined such that, for each element $x$ of a reference set $\referenceset$, it expresses the probability that $\drs$ contains $x$,
\begin{equation}
p_x(\drs) = \prob{x \in \drs}.
\end{equation}
%

\subsection{Distances}

Definitions of distances between points and sets, or sets and sets, commonly build on definitions of a distance measure between points. The Euclidean point-to-point distance $\deuclidean$ is a natural and common choice.

Given a point-to-point distance $d$, the standard point-to-set distance between a point $x$ and a set $X$ is defined as
\begin{equation}
\label{eq:crisppointtosetdist}
d(x, X) = \inf\limits_{y \in X}{d(x, y)}.
\end{equation}

The (unsigned, external) distance transform (DT) with respect to a foreground set (image object) $X$ evaluated on the image domain $I$, is 
\begin{equation}
\texttt{DT}[X](x) = \min\limits_{y \in X}{d(x, y)}, \qquad \text{for} \quad x \in I.
\end{equation}
Due to the separability of the (squared) Euclidean distance, the DT with $\deuclidean$ as underlying point-to-point distance can be computed exactly, in time linear to the number of pixels \cite{maurer2003linear}, which enables its efficient use in practical applications. 

Point-to-set distances can also be extended to DRS \cite{baddeley1998averaging} and hence generate probability distributions of distance values. Given a set of  realizations of a DRS $\drs_1, \drs_2, \drs_3 \ldots \drs_n$, 
an empirical mean (or some other statistics) of the distances $d(x,\drs_1), \ldots, d(x,\drs_n)$ can be computed, 
\cite{lewis1999averaging}, to estimate a distance of a given point $x$ to the DRS.

\section{Stochastic Distance Transform}

In this section we propose a novel type of noise-resistant DT, defined for (ordinary, non-random) 
sets and points. 
The transform builds on the theoretical framework of random sets, and distributions of distances from points to random sets, to achieve high robustness to noise.

Let $\randomize(\referenceset, c)$ denote a DRS on a reference set $\referenceset$, where probability of inclusion/exclusion of each element is i.i.d., that is,  independent from the inclusion/exclusion of all other elements, and identically distributed, with constant coverage probability $c$, 
i.e., let $p_x(\randomize(\referenceset, c))=c$, \,$\forall~x\in\referenceset$.

\begin{definition}
Given an image domain $I$, a foreground set (image object) $X \subseteq{I}$, uncertainty factor $\uncertaintyfactor \in \left[0, 1\right]$, a maximal distance $\dmax \in \mathbb{R}_{+}$, and a point-to-set distance $d$, the (unsigned, external) {\it Stochastic Distance Transform (SDT)} is 
\begin{equation}
\sdt{\uncertaintyfactor}{X}{x} = \EX(\min\left[d(x, \randomize(X, 1-\uncertaintyfactor)), \dmax\right]), \quad \text{for} \; \; x\in I.
\end{equation}
\end{definition}

For  $\uncertaintyfactor \in (0, 1]$,  there is a non-zero probability that all points from $X$ are excluded in some realization of $\randomize(X, 1-\uncertaintyfactor)$. Since $d(x, \emptyset) = \infty$ 
this leads to 
the expectation value $\EX(d(x, X)) = \infty$, for all $X$ and $x$.
Special care has, therefore,  to be taken of the case of empty sets, to ensure that the SDT is well defined. 
We propose one possible solution by introduction of a parameter, $\dmax$, a finite maximum distance which saturates the underlying point-to-set distance. This ensures that $\sdt{\uncertaintyfactor}{X}{x}$ is finite-valued and well-defined for all $\uncertaintyfactor \in [0, 1]$, $X$ and $x$.

Tuning of $\uncertaintyfactor$  depends on the amount of noise and artifacts in the images of interest, and is either performed by heuristics, or by optimisation of an application specific evaluation metric. $\dmax$ is typically set to the diameter of the domain.

\subsection{Monte Carlo Method}

An estimate of \sdt{\uncertaintyfactor}{X}{x} can be obtained by a Monte Carlo method, denoted MC-SDT, by drawing $N$ random samples (sets) from $R(\referenceset, 1-\uncertaintyfactor)$, computing the corresponding point-to-set distances, typically using a fast DT algorithm, and then computing their empirical mean:
\begin{equation}
\mcsdt{\uncertaintyfactor,N}{X}{x} = \frac{1}{N} \sum\limits_{i=1}^{N}{\min\left[d(x, R(X, 1-\uncertaintyfactor)_i), \dmax\right]}\,.
\end{equation}
Here $R(\referenceset, 1-\uncertaintyfactor)_i$ denotes realization $i$ of random set $R(X, 1-\uncertaintyfactor)$, which can be sampled by one independent Bernouilli trial per element in $\referenceset$. 

\subsection{Deterministic Method}

The $\sdt{\uncertaintyfactor}{X}{x}$ can be modeled similarly to a geometric distribution, where each trial has a corresponding distance. The nearest point to $x$ in $X$ will be present and selected with probability $1-\uncertaintyfactor$; the second nearest point to $x$ in $X$ will be present and selected with probability $\uncertaintyfactor(1-\uncertaintyfactor)$, and hence the $i$-th nearest point in $X$ will be present and selected with probability $\uncertaintyfactor^{\,i-1}(1-\uncertaintyfactor)$, given that such a point exists. 

Let 
$d_{(i)}(x, X)$ denote a generalization of the point-to-set distance \eqref{eq:crisppointtosetdist}, which defines the distance between the point $x$ and the set $X$ to be the distance between $x$ and its  $i$-th nearest point in $X$, where $d_{(i)}(x, X) = \infty, \;\text{for}\; i > |X|$.
Now, a deterministic formulation of SDT, denoted DET-SDT can be given as:
\begin{equation}
\label{eq:detsdt}
\detsdt{\uncertaintyfactor, k}{X}{x} = \uncertaintyfactor^k \dmax + \sum\limits_{i = 1}^{k}{\rho^{i-1}(1-\rho) \min \left[ d_{(i)}(x, X), \dmax \right]},
\end{equation}
where $k$ denotes the number of considered nearest points, and $0^0=1$.

The DET-SDT method given in \eqref{eq:detsdt} is exactly equal to the SDT if $k=|X|$, i.e. all points in $X$ are considered. In practice it tends to be impractical to consider all the points in $X$ (especially considering the exponentially diminishing contribution of each additional point) and we may instead choose to capture a sufficiently large fraction $m$ (for the application at hand) of the probability mass, such as $m=0.999$. Given such a value $m \in (0, 1)$ and $\uncertaintyfactor > 0$, using the cumulative distribution function (CDF) of the geometric distribution, we can solve for an integer $\kappa_{\uncertaintyfactor, m}$ of minimally required nearest points which  guarantee that at least $m$ of the total probability mass is captured,
\begin{equation}
\label{eq:kappa}
\kappa_{\uncertaintyfactor, m} = \Bigl\lceil \frac{log(1-m)}{log(\uncertaintyfactor)} \Bigr\rceil.
\end{equation}
Table \ref{tab:kappa} shows the required number of points to consider for various $m$ and $\uncertaintyfactor$. 
\begin{table}[ht]
\centering
\caption{Minimally required number of nearest points $\kappa_{\uncertaintyfactor, m}$ to consider for various combinations of probability mass $m$ and uncertainty factor $\uncertaintyfactor$.}
\label{tab:kappa}
\begin{tabular}{c@{\;}|@{\;}cccccccccccc}
m & \multicolumn{12}{c}{$\uncertaintyfactor$} \\ \hline
& 0.1 & 0.2 & 0.3 & 0.4 & 0.5 & 0.6 & 0.7 & 0.8 & 0.9 & 0.95 & 0.975 & 0.99 \\ \hline
$0.95$ & 2 & 2 & 3 & 4 & 5 & 6 & 9 & 14 & 29 & 59 & 119 & 299 \\ 
$0.99$ & 2 & 3 & 4 & 6 & 7 & 10 & 13 & 21 & 44 & 90 & 182 & 459 \\ 
$0.999$ & 3 & 5 & 6 & 8 & 10 & 14 & 20 & 31 & 66 & 135 & 273 & 688 \\ 
\end{tabular}
\end{table}

There are many algorithms in the literature for finding the $k$-nearest neighbours (k-NN) among a set of points, with corresponding distances. For regularly spaced grids, there are efficient algorithms for computing the k-NN utilizing the properties of the grid to achieve an improved computational complexity \cite{cuisenaire2000fast}. For other scenarios, e.g. for point-clouds, algorithms based on the efficient kd-tree data-structure \cite{bentley1975multidimensional} can be used to compute the k-NN efficiently. Once the k-NN (with distances) has been found, the closed-form expression \eqref{eq:detsdt} can be computed directly. The best algorithm and data-structures for computing the k-NN for \eqref{eq:detsdt} is highly situation-dependent, and a trade-off must be found between factors such as: (i) execution time; (ii) memory usage; (iii) utilization of the image domain structure.  

\section{Performance Analysis}

In this section we evaluate the utility and performance of the proposed method in three main ways: (i) Measuring the distance accuracy in the presence of noise; (ii) Measuring the effect of the SDT on robustness of a template matching framework, when the proposed method is inserted 
into a set-to-set measure based on spatial/shape information; (iii) Observing the difference in quality of the segmentation obtained by replacing the standard DT with the SDT in the context of the classical watershed segmentation framework.

If not stated otherwise in experimental setups, $N=400$ realizations are used for the MC-SDT, and at least $m=99.9 \%$ of the probability mass is used for the DET-SDT. The parameter $k$ is  determined by this $m$ and the used $\uncertaintyfactor$, according to \eqref{eq:kappa}, and as illustrated in Tab.~\ref{tab:kappa}.

\subsection{Distance Transform Accuracy in the Presence of Noise}
\label{sec:acc}

The accuracy of the distances computed by the standard DT can deteriorate heavily with just a single noise-point in a background region. In this section, the accuracy of the SDT is compared to the accuracy of DT,  in the presence of added noise.

\paragraph{Experimental Setup:}

\begin{figure}[ht]
\centering
\begin{tabular}[t]{ccccc}
\includegraphics[width=2.55cm]{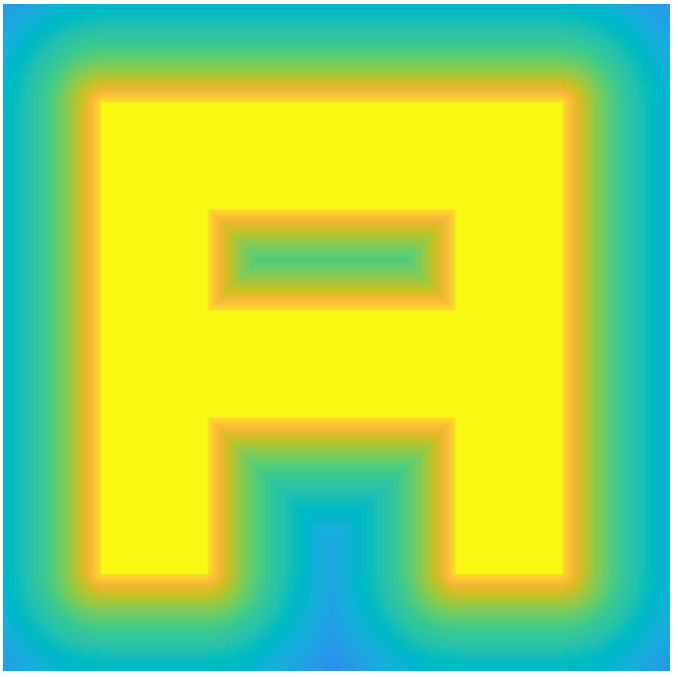} &
\includegraphics[width=2.55cm]{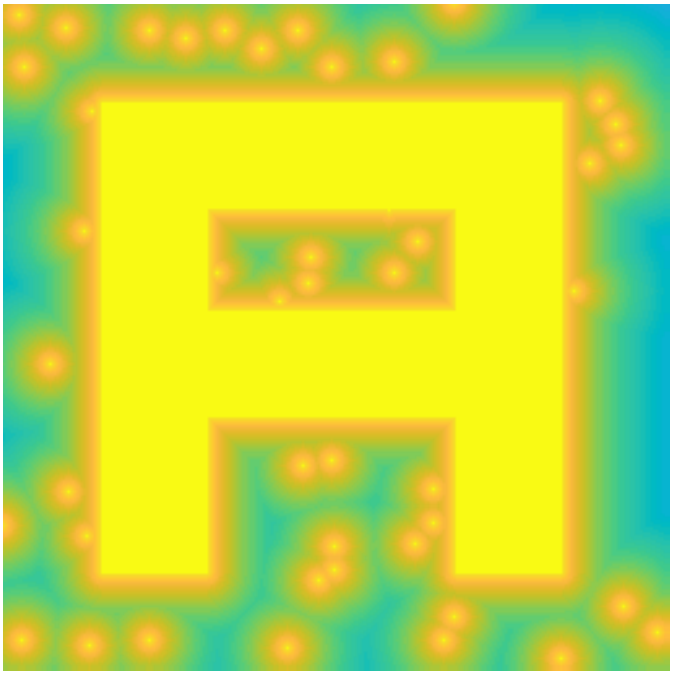} &
\includegraphics[width=2.55cm]{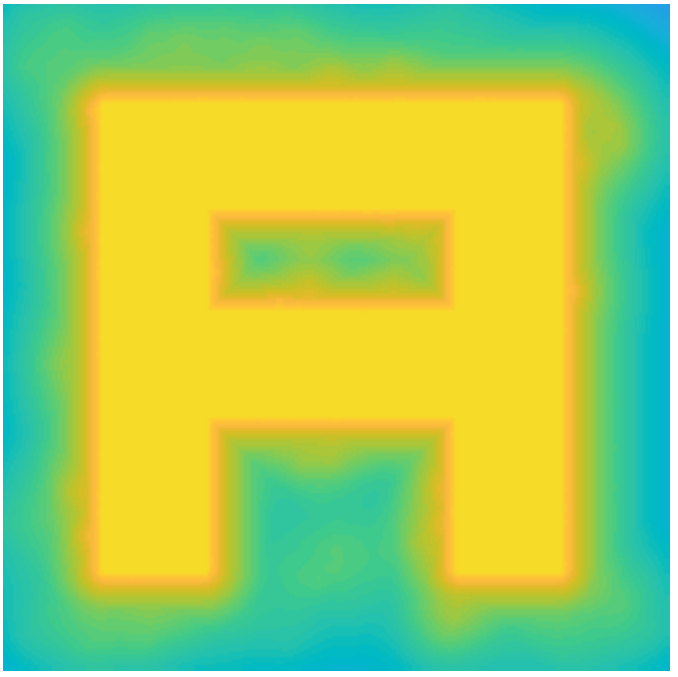} &
\includegraphics[width=2.55cm]{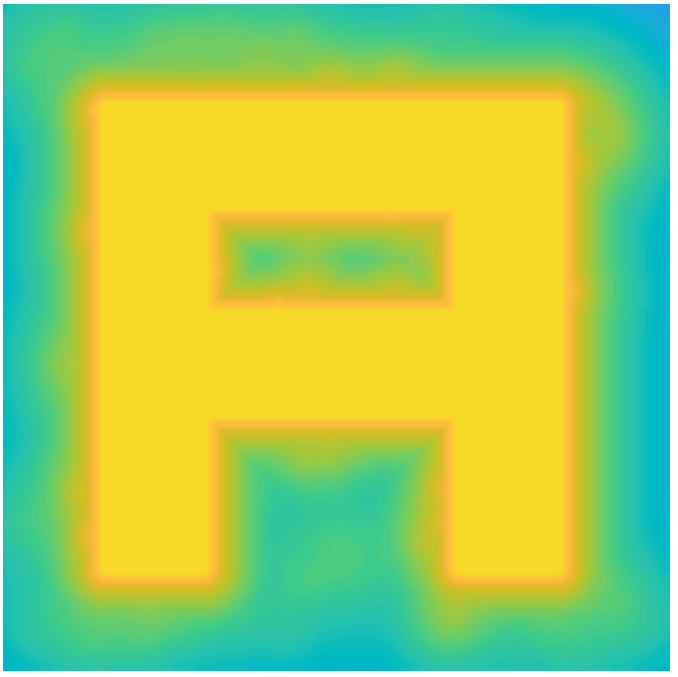} &
\multirow{4}{*}[2cm]{
\begin{tabular}{c}
$d_\texttt{MAX}$ \\
\includegraphics[height=4.5cm]{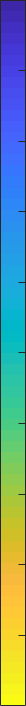} \\
0
\end{tabular}
} \\
(a) Noise-free DT & (b) DT & (c) MC-SDT & (d) DET-SDT & \\
\includegraphics[width=2.55cm]{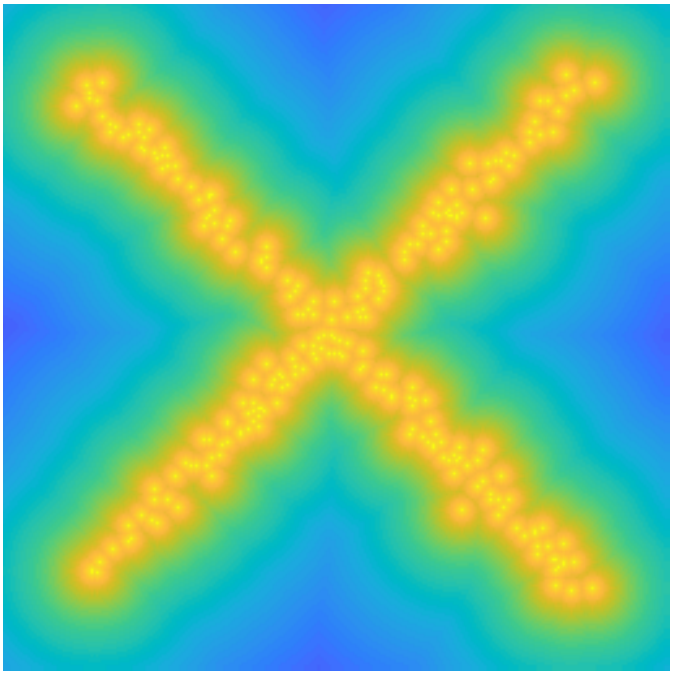} &
\includegraphics[width=2.55cm]{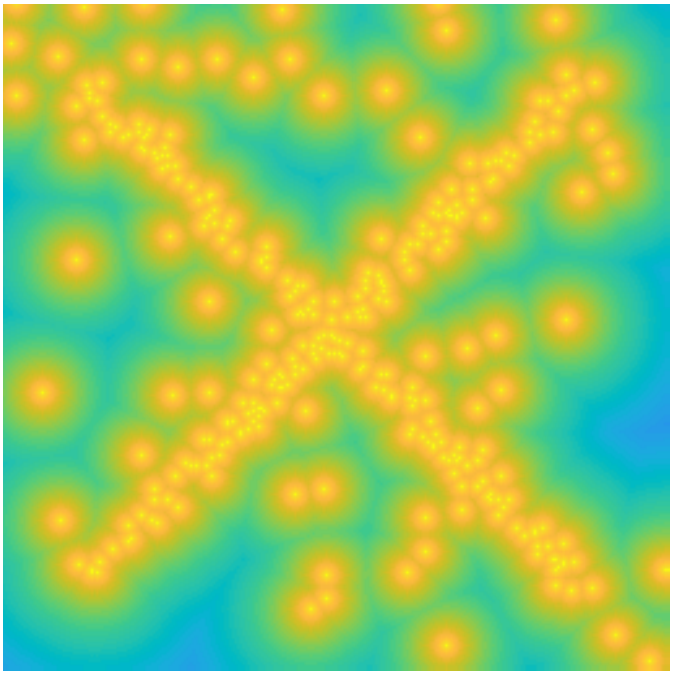} &
\includegraphics[width=2.55cm]{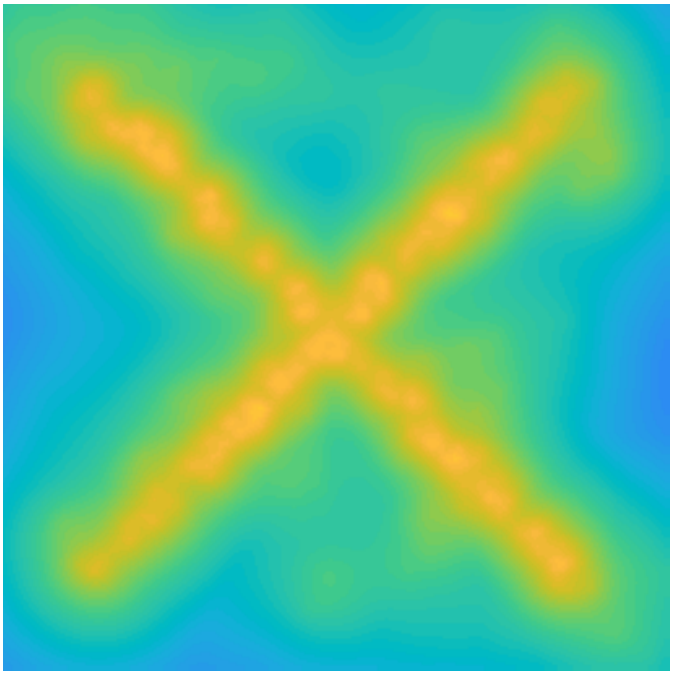} &
\includegraphics[width=2.55cm]{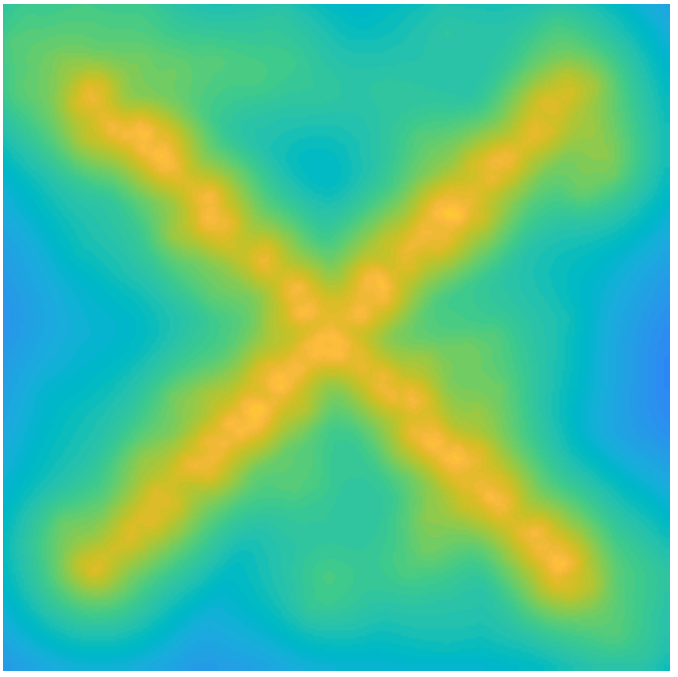} &
 \\
(e) Noise-free DT & (f) DT & (g) MC-SDT & (h) DET-SDT & \\
\end{tabular}

\caption{(a,e) Noise-free DT of the test images; (b,f) DT of the same images, after they are corrupted by noise; (c,d,g,h) SDT applied on the noisy images. }
\label{fig:letters}
\end{figure}

We consider two test images, one containing a solid letter A, and the other containing a letter X constructed as a sparse point-cloud in the regular grid, both corrupted by random noise-points  added with probability $p=0.001$. We compute both the MC-SDT and DET-SDT  using $\uncertaintyfactor = 0.75$. Different computed DTs, in noise-free and noisy conditions, are presented  in Fig.~\ref{fig:letters} for qualitative assessment. The evaluation metric used is Average Absolute Distance Error (AADE) in the computed distance map, over all pixels, averaged over 100 repetitions with different noise realizations. 

\paragraph{Results:}

Quantitative results are presented in Tab. \ref{tab:aade}. The stochastic methods exhibit substantially higher, and more consistent (in terms of std. dev.), accuracy than the standard DT in the presence of noise.
 
\begin{table}[ht]
\caption{Average absolute distance error (AADE \rpm SD) for the experiments illustrated in Fig.~\ref{fig:letters}. Lower is better. Bold marks the best result for each image.}
\label{tab:aade}
\centering
\vspace{0.1cm}
\begin{tabular}{@{\;}l@{\;}|@{\;}c@{\;}|@{\;}c@{\;}|@{\;}c@{\;}}
& DT & MC-SDT & DET-SDT \\ \hline
Image A \Tstrut[2.4ex] \hspace{0.1cm} & \,5.41 \rpm 0.55 & \textbf{2.39} \rpm 0.29 & \textbf{2.42} \rpm 0.27 \\
Image X \hspace{0.1cm} & 14.31 \rpm 0.87 & \textbf{8.02} \rpm 0.75 & \textbf{8.02} \rpm 0.74 \\
\end{tabular}
\end{table}



\subsection{Template Matching}

Template matching of (binary) images is a process of locating a particular region/object in the image by finding a location where a given template "fits best", i.e., where 
a distance between the template and the image is minimal (or where a similarity measure is maximal). 
In the search, we consider all possible translations of the observed template 
by vectors with integer coordinates, such that the template is completely included in the image. 
We minimize the bidirectional (asymmetric, being suitable for template matching) distance \cite{lindblad2009set} based on \emph{Sum of Minimal Distances} (SMD) \cite{eiter1997distance}, defined as
\begin{equation}
    \darrow(A, B) = \sum\limits_{a \in A}{d(a, B)} + \sum\limits_{a \in \bar{A}}{d(a, \bar{B})},
\end{equation}
where $\bar{A}$ and $\bar{B}$ denote the complement sets of $A$ and $B$, respectively.
This distance measure
has been shown to have a number of appealing properties, such as a smooth distance field subject to translation, rotation and affine transformations. One drawback that has been observed is that this distance is quite noise-sensitive in the sense that a few spurious points can create shallow local minima in its distance landscape. As a consequence, both local search (where the search stops upon finding a local minimum) or global search may result in   
many false detections
and must be pruned in post-processing. This part of the study aims to investigate if noise-sensitivity of template matching with $\darrow$ can be reduced if SDT is used in  computation of $\darrow$ instead of the (ordinary) DT.

\paragraph{Experimental Setup:}

We consider 
the  well-known \textbf{Cameraman} (grey-scale) image. We corrupt it with additive Gaussian noise ($\sigma=0.1$), Fig. \ref{fig:tmresults1}(a) and then threshold at intensity $0.5$ into a binary image, Fig. \ref{fig:tmresults1}(b). A binary template is extracted from the noise-free original, by thresholding at the same intensity level, Fig. \ref{fig:tmresults1}(c). Within the evaluation framework we compute the distance $\darrow(T, X)$ between the template $T$  and the image $X$, for every position in the image where the template is completely included in the image. In this computation, we use $SDT_{\uncertaintyfactor}$, with $\uncertaintyfactor \in \left\{ 0, 0.025, 0.05, \ldots 0.975, 0.99 \right\}$, as the underlying DT for $\darrow$.  The position where global minimum of $\darrow$ is reached is recorded to evaluate if the correct location is recovered. The number of minima (NoM) is also computed for the distance field, as well as the catchment basin (CB) of the global minimum. The CB is the set of all image points which would, if used as initialization for a local search (using 8-neighbourhood steps), provide convergence to the global minimum. The evaluation metrics are averaged over $50$ noise realizations for each considered $\uncertaintyfactor$.
Since the uncertainty factor $\uncertaintyfactor=0$ corresponds to the standard DT, this evaluation includes comparison of performance of $\darrow$ using standard DT and with using the here proposed SDT.

\paragraph{Results:}

The results are presented in Fig. \ref{fig:tmresults1}(d-g). Figures \ref{fig:tmresults1}(d,e) show colored labelling (on a single realization) of the NoM and their corresponding CB, for standard DT, and for DET-SDT. Significantly decreased NoM, and visibly larger CB corresponding to the correct template position (red cross) characterize DET-SDT. The plots \ref{fig:tmresults1}(f,g) show the NoM and the size of a CB of the global minimum (in a percentage of the number of pixels in the image), as a function of the uncertainty factor $\uncertaintyfactor$. The results clearly show that the evaluation metrics improve in a stable and gradual way with increasing $\uncertaintyfactor$. The NoM exhibits a linearly decreasing trend, where the CB size initially exhibits a linearly increasing trend until $\uncertaintyfactor > 0.7$, where a super-linear increase is observed. The know global minimum (correct match) is successfully recovered in all tests.

\afterpage{
\begin{figure}[ht]
\centering
\subfloat[][Noisy Input Image]{\includegraphics[width=2.3cm]{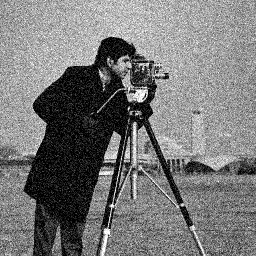}}\hspace{\fill}
\subfloat[][Binary Image]{\includegraphics[width=2.3cm]{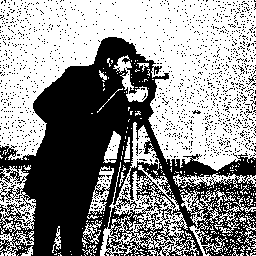}}\hspace{\fill}
\subfloat[][Binary Template]{\includegraphics[width=2cm]{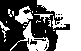}}\hspace{\fill}
\subfloat[][DT: CB Labelling]{\includegraphics[width=2.3cm]{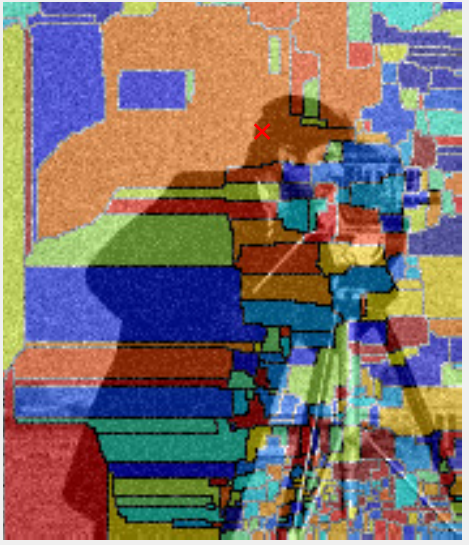}}\hspace{\fill}
\subfloat[][DET-SDT: CB Labelling]{\includegraphics[width=2.3cm]{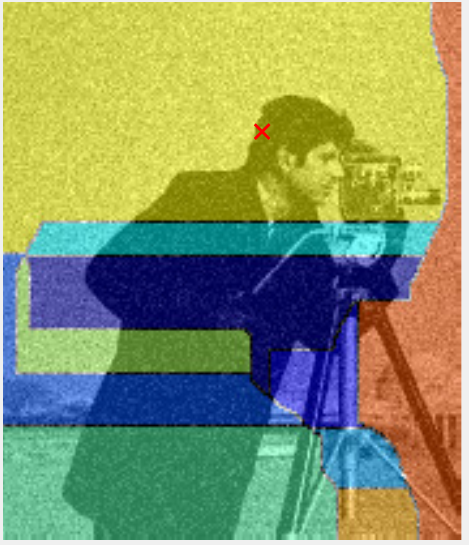}}\\
\subfloat[][Number of minima (NoM)]{\includegraphics[width=5cm]{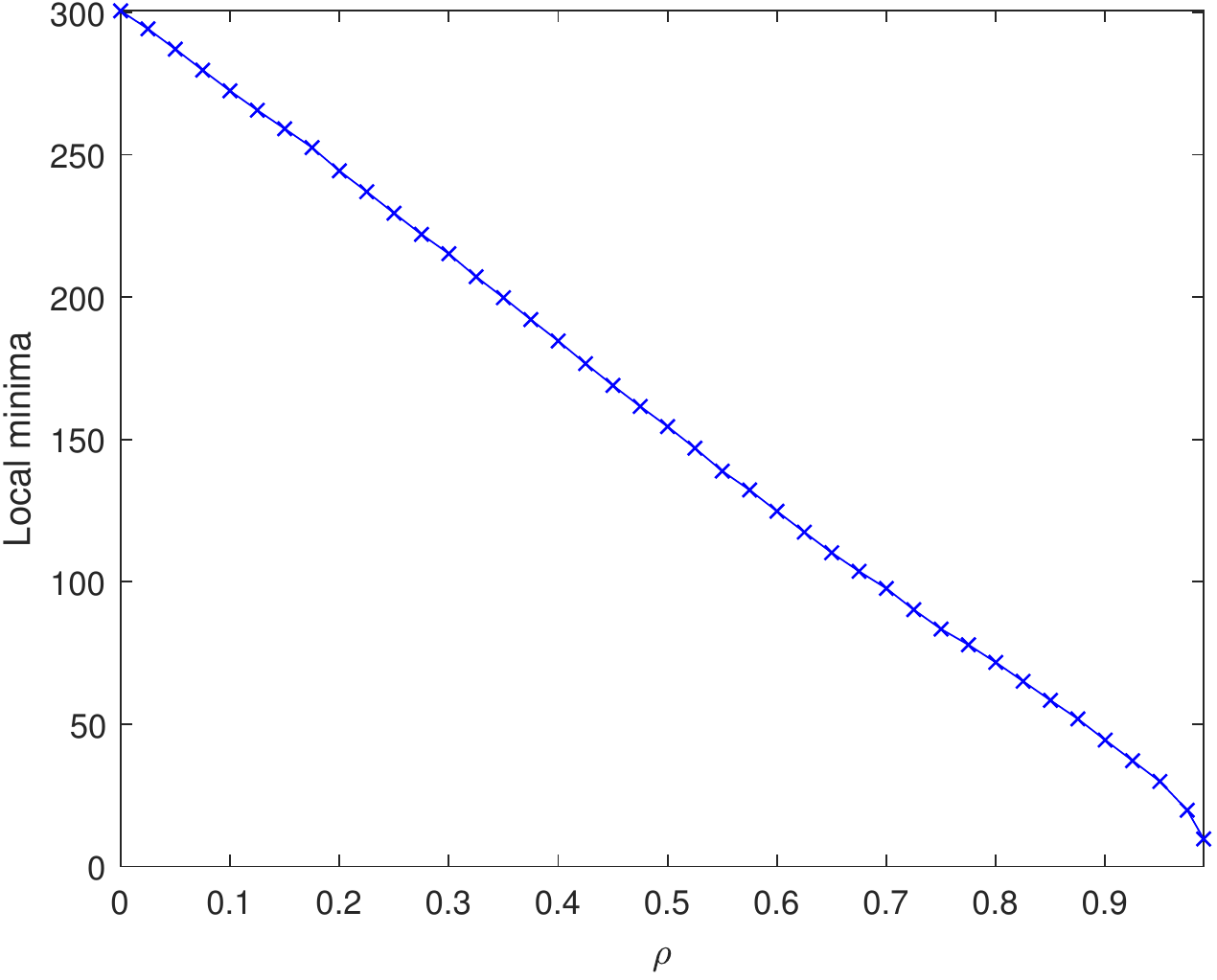}}\hspace{\fill}
\subfloat[][Global Minimum CB Size]{\includegraphics[width=5cm]{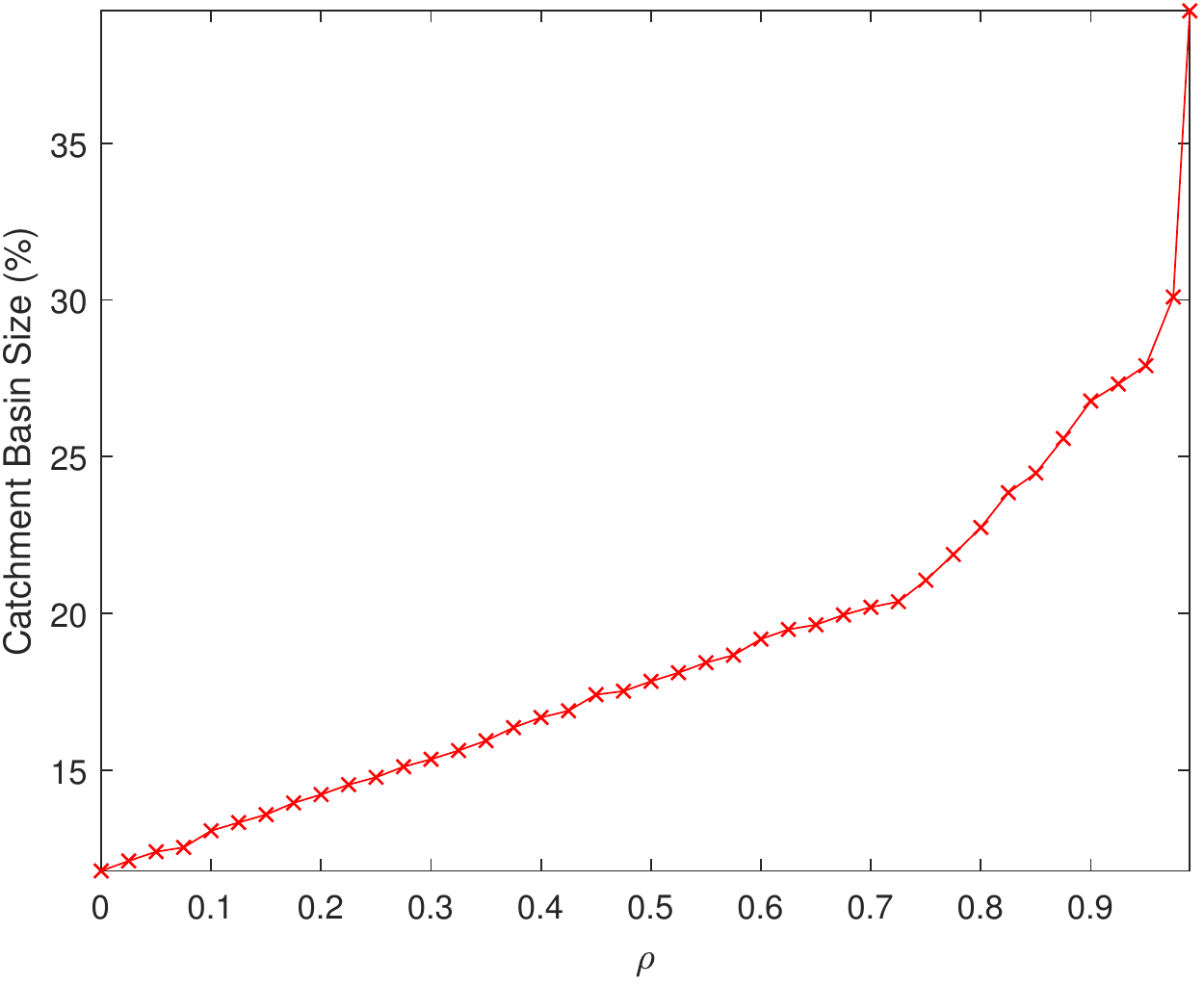}}\hspace{\fill}\\
\caption{Template matching on binarized versions (b) of a test image (a) with additive Gaussian noise ($\sigma=0.1$), and a noise-free template (c). The labelled catchment basins (CB) for each local minimum with standard DT (d), and $\texttt{SDT}_{0.99}$ (e). 
NoM (f) and size of CB (g) for different  values of $\uncertaintyfactor$ used in the DET-SDT. The MC-SDT exhibits very similar performance in this experiment.}
\label{fig:tmresults1}
\end{figure}
}

\subsection{Watershed Segmentation}

The watershed transform \cite{beucher1979use} is a transformation which partitions a grey-scale image into regions associated with 
the local minima of the image (or a number of defined seed points).
Intuitively, the graph of the grey-level image is flooded with water coming out from the seed points (minima) and filling the corresponding basins. Where the basins of different seed points meet, ridge-lines mark a delineation of the different objects. 
One common approach for shape based watershed segmentation is to use the negative of the DT as the grey-scale image, and its minima (maxima in the original DT) as seed-points. 

It is important to note that the watershed segmentation approach used here utilizes stochasticity in a very different way than the stochastic watershed segmentation \cite{angulo2007stochastic} method, which randomly places seed points and yields a PDF of the ridge-lines which separate the objects in the image. The  method presented here employs stochasticity to remove spurious optima, with the aim of achieving robustness to noise and preventing oversegmentation.

\subsubsection{Separation of a Pair of Discretized Disks}

\paragraph{Experimental Setup:}

\afterpage{
\begin{figure}[ht]
\centering
\hspace{\fill}\subfloat[][DT]{\includegraphics[width = 2.5cm,trim=1cm 1cm 1cm 1cm,clip]{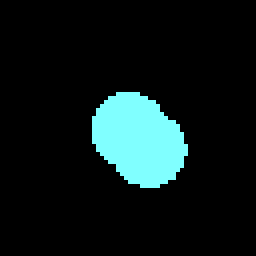}} \hspace{\fill}
\subfloat[][MC-SDT]{\includegraphics[width = 2.5cm,trim=1cm 1cm 1cm 1cm,clip]{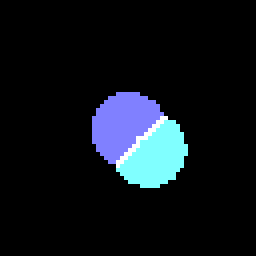}}\hspace{\fill}
\subfloat[][DET-SDT]{\includegraphics[width = 2.5cm,trim=1cm 1cm 1cm 1cm,clip]{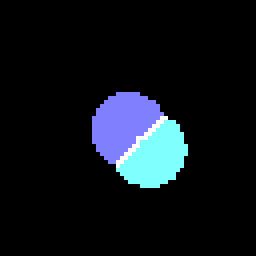}}
\hspace{\fill}\,\\
\vspace{0.5cm}
\subfloat[][DT]{\includegraphics[width=12cm]{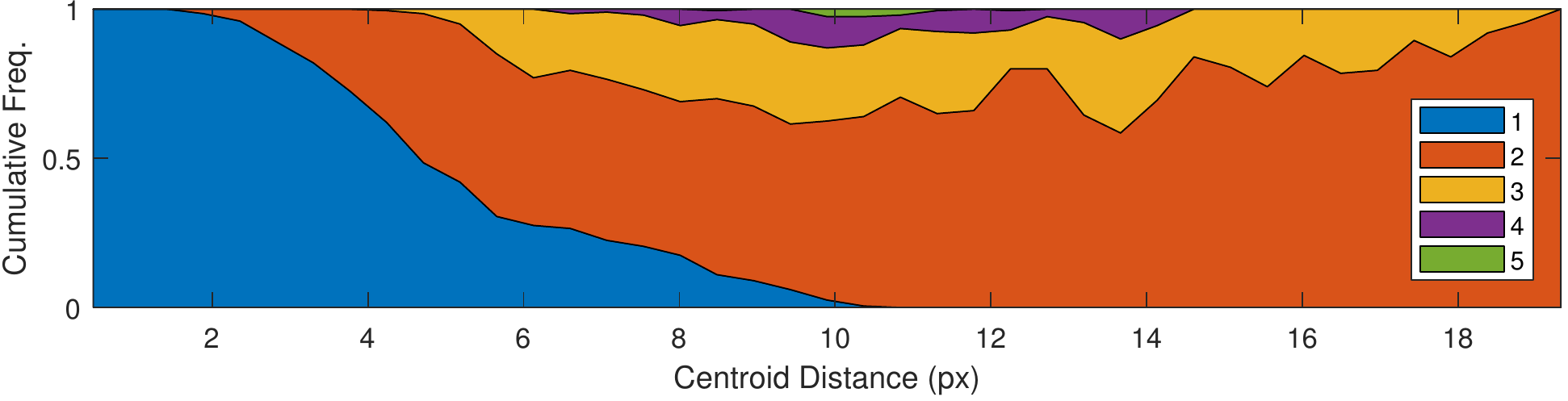}}\hspace{\fill}\\
\subfloat[][MC-SDT]{\includegraphics[width=12cm]{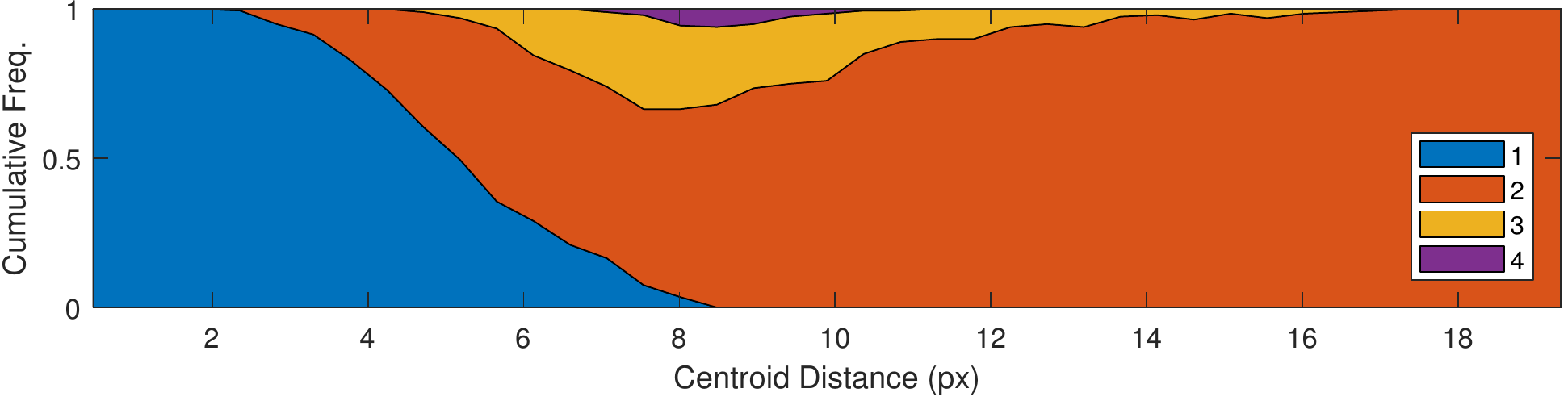}}\hspace{\fill}\\
\subfloat[][DET-SDT]{\includegraphics[width=12cm]{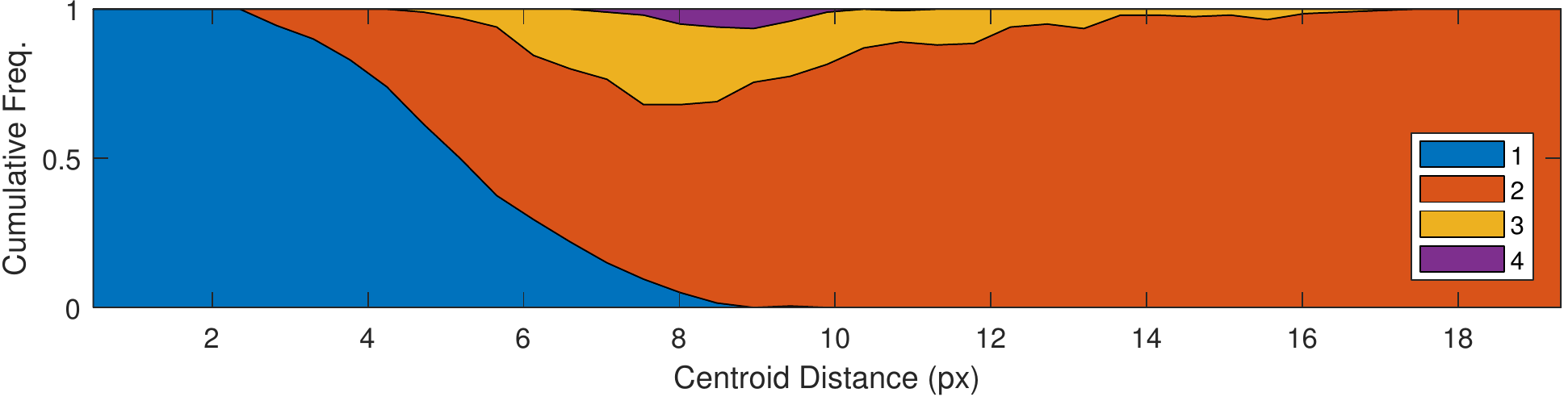}}
\caption{(a-c) An example of disks segmentation by watershed algorithm based on different distance transforms: DT, MC-SDT, and DET-SDT. Segmentation based on the standard DT resulted in a single object, while segmentations with both SDT approaches successfully separate the two objects. (d-f) Quantitative evaluation of the performance on disk separation. The frequencies of appearance of the different segment counts are presented as colored areas, for increasing distance between the centers of the disks.  The improvement brought by SDT, over DT, is indicated by absence of 5-segment results, and larger number of 2-segment results (indicated by the larger corresponding area), than for DT.}
\label{fig:wsresults1}
\end{figure}
\clearpage
}

To evaluate the proposed method in a scenario with no additive noise, but merely digitization artifacts/noise, a synthetic benchmark was constructed: two equisized disks are positioned  with random sub-pixel placements, so that they have some overlap, and then digitized (by Gauss centre point digitisation) on a regular grid into a binary image. Watershed segmentation is applied on the negated internal DT to separate the created object into two components. 
Figure~\ref{fig:wsresults1}(a) shows the created binary object, where (b) an (c) present  examples of segmentation (separation of the two disks). The radii of the disks are chosen to be $r = 3 \pi$ pixels, to create reasonably sized objects with irrational radii. Following digitization, watershed segmentation is applied to the distance map resulting from the DT, MC-SDT and DET-SDT on the binary image. The resulting segmentations are analyzed w.r.t. the number of segmented objects, observing 200 repetitions of the experiment for every value of the distance between the centres of the disk within the range $[0.05r, 2r]$, with a step-size of $0.05r$.  
Considering that the disks can, in the continuous case, always be segmented into two components, 
we assume that 2 is the correct number of objects to result from the performed watershed segmentation. 
 An uncertainty factor $\uncertaintyfactor=0.75$ is selected based on tuning on a smaller set of repetitions and steps.


\paragraph{Results:}

Figure \ref{fig:wsresults1} shows the results of the disks segmentation (separation) experiment. Each plot shows the performance of the segmentation in terms of the fraction of the trials at which the various object counts occur, as a function of distance between the disk centres. We observe that the SDT-based methods perform similarly, and substantially better than the standard DT. Table \ref{tab:wsresults1} shows the Area Under the Curve (AUC) of the detection frequency corresponding to 2 segments.

\afterpage{
\begin{figure}[ht]
\centering
\hspace{\fill}
\subfloat[][Input Image]{\includegraphics[width=5cm]{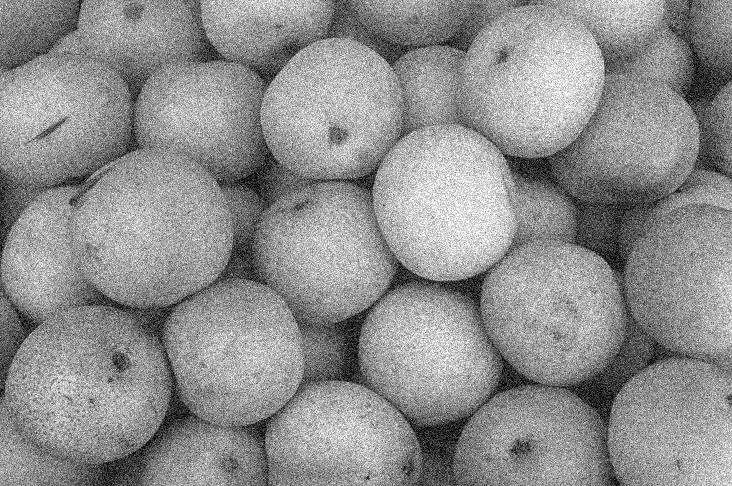}} \hspace{\fill}
\subfloat[][Binary Image]{\includegraphics[width=5cm]{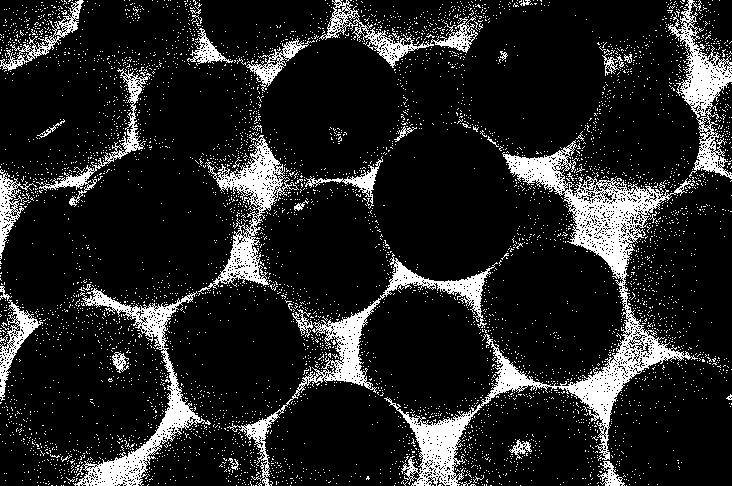}}
\hspace{\fill}\,\\
\subfloat[][DT: Overlay]{\includegraphics[width=3.9cm]{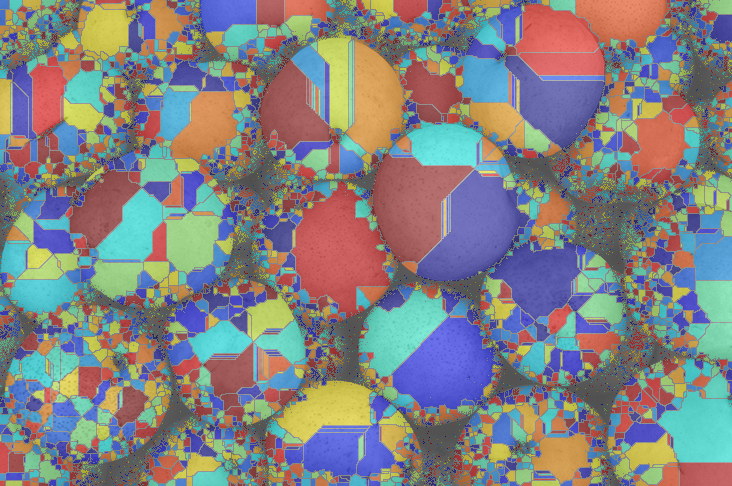}}\hspace{\fill}
\subfloat[][MC-SDT: Overlay]{\includegraphics[width=3.9cm]{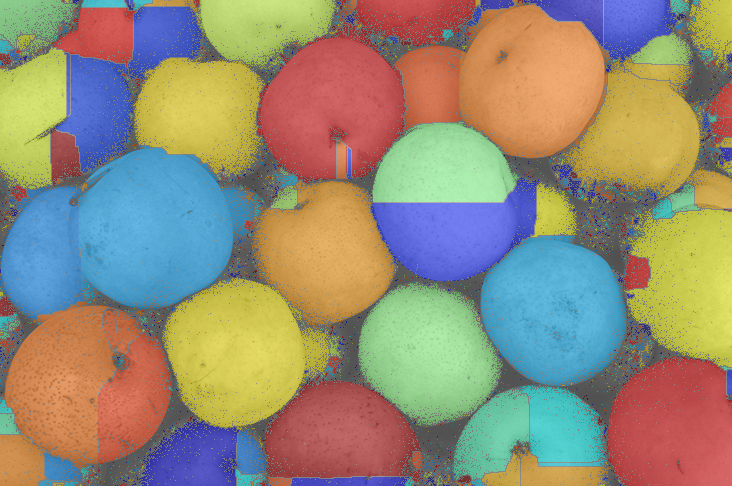}}\hspace{\fill}
\subfloat[][DET-SDT: Overlay]{\includegraphics[width=3.9cm]{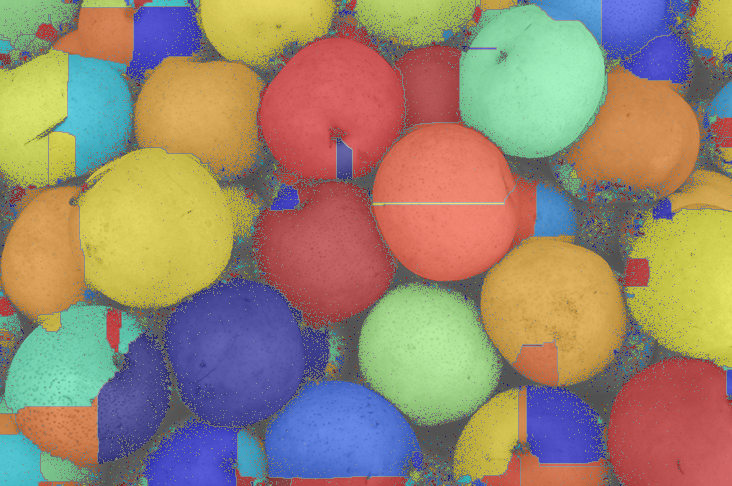}}
\caption{(a) An image corrupted by a moderate amount of Gaussian noise ($\sigma=0.1$); (b) Binarization of (a) by thresholding;  (c-e) The labellings obtained by watershed segmentation with the standard DT, MC-SDT, and DET-SDT, respectively, overlayed on the image. Using the classic DT yields a highly over-segmented image while both variations of the SDT yield segmentations that are largely unaffected by the noise.}
\label{fig:wsresults2}
\end{figure}
}

\begin{table}[ht]
\centering
\caption{Disk separation by watershed segmentation (Fig. \ref{fig:wsresults1}): AUC measure for the detection frequency corresponding to 2 segments. Higher is better. The best result is presented in bold.}
\label{tab:wsresults1}
\begin{tabular}{|@{\;}r@{\;}||@{\;}c@{\;}|@{\;}c@{\;}|@{\;}c@{\;}|}
    \hline 
    Method & DT & MC-SDT & DET-SDT  \\ \hline
    AUC: (2 segments) & 0.563 & 0.661 & $\mathbf{0.663}$ \\ \hline
\end{tabular}
\end{table}

\subsubsection{Watershed segmentation, a realistic example}

\paragraph{Experimental Setup:}

To evaluate the performance of the watershed segmentation  when used with the proposed SDT method in a realistic setting, we observe the well known image {\bf Pears.png}, to which we apply additive Gaussian noise with $\sigma = 0.1$,  Fig. \ref{fig:wsresults2}(a). We  binarize the image (by thresholding),  Fig. \ref{fig:wsresults2}(b), and segment it using watershed method with both DT and SDT.    

Parameter values are: $\uncertaintyfactor = 0.95$, $\dmax = 256$, binarization threshold set to $0.35$ (manually selected on the noise-less image).

\paragraph{Results:}

The segmentation results are evaluated subjectively. We find that the segmentations shown in  Fig. \ref{fig:wsresults2}(d,e), which rely on SDT, clearly indicate advantage of the here presented approach, compared to classic DT  which leads to heavy oversegmentation, caused by high sensitivity to noise. 

\section{Conclusion}

In this study we have proposed a novel type of distance transform, the Stochastic Distance Transform. SDT is based on probability distributions of distances to image objects represented as Discrete Random Sets. The main advantage of the SDT over the classic DT is its adjustable robustness to noise, allowing to choose parameters controlling a level of sensitivity according to the application at hand. The proposed method's utility and favorable properties are observed both through various synthetic tests and an illustrative natural example.

Future work includes an extended study of the theoretical properties of the proposed method, investigating (possibly adaptive) methods for reducing the biases of the resulting distance values, extending the empirical evaluation, and exploring further potential applications.

\section*{Acknowledgements}

This work is supported by VINNOVA, MedTech4Health grants 2016-02329 and
2017-02447 and the Ministry of Education, Science, and Techn. Development of the
Republic  of  Serbia  (proj.  ON174008  and  III44006).

%
%
%
\bibliographystyle{splncs04}
\bibliography{references}
%





\end{document}